\documentclass{bmvc2k}

\usepackage{graphicx}
\usepackage{tabularx}
\usepackage{array}
\usepackage{booktabs}
\usepackage{placeins}
\usepackage{url}

\usepackage[table]{xcolor}
\usepackage{makecell}

\definecolor{HeaderBlue}{RGB}{226,237,247}
\definecolor{GroupGray}{RGB}{245,245,245}
\definecolor{BestGreen}{RGB}{213,236,204}
\definecolor{SecondYellow}{RGB}{250,239,195}

\newcolumntype{Y}{>{\raggedright\arraybackslash}X}
\newcolumntype{L}[1]{>{\raggedright\arraybackslash}p{#1}}

\title{EgoMaize: A First-Person Maize Instance Segmentation Benchmark under Severe Field Occlusion}

\addauthor{Jiayi Li}{lijiayi12jad@gmail.com}{1}
\addauthor{Zihan Zhang\textsuperscript{*}}{ffgs20050628@gmail.com}{1}
\addauthor{Erhankang Yan\textsuperscript{*}}{khangy4n@gmail.com}{2}
\addauthor{Yitian Chen\textsuperscript{*}}{chenyitian0607@outlook.com}{1}
\addauthor{Yuze Li\textsuperscript{*}}{2932761898lyz@gmail.com}{1}
\addauthor{Chengzhang Ding}{ddaihh@emails.bjut.edu.cn}{1}
\addauthor{Jianxin Cao}{caojianxin@emails.bjut.edu.cn}{1}

\addinstitution{
College of Computer Science\\
Beijing University of Technology\\
Beijing, China
}

\addinstitution{
School of Mathematics, Statistics and Mechanics\\
Beijing University of Technology\\
Beijing, China
}

\runninghead{Li et al.}{EgoMaize}

\graphicspath{{images/}}

\begin{document}

\maketitle

\begingroup
\renewcommand{\thefootnote}{*}
\footnotetext{These authors contributed equally.}
\endgroup

\begin{abstract}
Close-range first-person field images are important for mobile maize
phenotyping because many plant-level traits depend on in-canopy structures
that are difficult to observe from overhead views. However, post-seedling
maize fields create a difficult instance segmentation setting: stems,
leaves, tassels, and neighboring plants are elongated, repetitive, and
strongly occluded. We introduce EgoMaize, a compact benchmark for
first-person maize instance segmentation, where the task is to predict
ownership-consistent plant masks and plant-owned stem/tassel cues from
close-range field images with severe same-class overlap. Existing
visible-only labels can fragment one physical plant into disconnected
supervision, while full-amodal labels may require unverifiable completion
behind neighboring plants or field objects. EgoMaize therefore uses an
evidence-closed annotation workflow for occluded maize regions and assigns
unreliable maize regions to ignore rather than background. Baseline
results show that pretrained query-based grouping, boundary refinement,
and high-resolution crop refinement help different aspects of the task,
but no architecture solves the coupled challenges of fine structure
recovery, same-class instance ownership, and occlusion reasoning;
occlusion-level analysis further shows that performance decreases as
plant visibility becomes more limited.
The dataset and code are publicly available at
\url{https://github.com/JaaaaaaaD/EgoMaize}.
\end{abstract}

\section{Introduction}
\label{sec:intro}

Close-range maize phenotyping requires perception of individual plants
from within the field, not only crop foreground detection. Many
agronomically relevant phenotypes are expressed in in-canopy structures:
stalk traits such as stem diameter and bending strength are used to
assess lodging resistance, with stalk lodging alone estimated to cause
5--20\% annual yield losses~\cite{Stubbs2022StalkGeometry}. Recent
robotic phenotyping studies further connect in-canopy traits such as ear
height, plant height, leaf area index, and stem diameter with biomass,
nitrogen response, and grain-yield variation
~\cite{Fan2022StemDiameterRobot,debruin2025maizeRobot}.
These traits depend on stems, tassels, leaves, and their spatial
arrangement, which are often difficult to observe from overhead views
alone. Extracting such traits from field images therefore first requires
reliable maize segmentation; for plant-level phenotyping, segmentation
must further assign stems, tassels, leaves, and fragmented visible
regions to the correct individual plant. In maize phenotyping, UAVs and
above-canopy sensing are effective
for large-scale traits such as canopy dynamics, vegetation indices, and
plant height, but in-canopy traits such as ear height, stem diameter,
organ arrangement, and plant-level architecture remain difficult to
obtain from above-canopy images~\cite{Han2018MaizeHTP,
Wang2023MaizeEarHeight,debruin2025maizeRobot}. Field robots and mobile
phenotyping platforms therefore increasingly observe crops from within
rows or canopies to automate fine-scale field phenotyping that is
labor-intensive or inefficient to perform manually
~\cite{Ninomiya2022FieldHTP,xu2022groundrobotreview,
debruin2025maizeRobot}.

This plant-level requirement is not fully covered by existing agricultural
vision benchmarks. Agricultural vision has benefited from crop--weed
segmentation datasets, UAV imagery, plant and leaf instance benchmarks,
and 3D phenotyping resources~\cite{weyler2024phenobench,
steininger2023cropandweed,celikkan2025weedsgalore,genze2024mfwd,
geng2024maizeseedling,schunck2021pheno4d,Zhu2024Crops3D}. These resources
support important tasks such as crop/weed recognition, semantic
segmentation, panoptic plant interpretation, organ-level analysis, and 3D
plant measurement. However, these benchmarks mainly address crop--weed
separation, overhead plant monitoring, seedling-stage instances,
organ-level labels, or 3D geometry. They do not provide a focused 2D
benchmark for close-range first-person post-seedling maize, where stems,
leaves, tassels, and fragmented visible regions from neighboring plants
must be assigned to the correct plant instance under severe same-class
occlusion. This gap matters because phenotyping and robotic monitoring
require stable association of plant parts, rather than only vegetation
coverage, crop/weed labels, or plant center detection. It therefore
motivates a dedicated benchmark that defines both the close-range
plant-ownership task and the annotation rules needed for severe field
occlusion.

EgoMaize is designed for this missing setting. It is a compact
first-person maize instance segmentation dataset for post-seedling field
scenes under severe same-class overlap. It contains 301 first-person field
images, 1,276 plant instances, and 2,731 ignore regions. Each plant
instance mask covers maize pixels that can be assigned to one physical
plant, while stem and tassel annotations are stored as plant-owned
auxiliary cues. We deliberately describe EgoMaize as compact rather than
large-scale: its value lies in focused annotation of a practical
agricultural engineering scenario where current instance segmentation
models expose clear failure modes.

Building such a benchmark raises a non-trivial annotation problem. In
post-seedling maize fields, plants are elongated, repetitive, and strongly
interleaved; stems, leaves, and tassels from neighboring plants often
cross or overlap in the image. If annotation keeps only visible pixels,
one physical plant may be split into several disconnected mask fragments.
If annotation instead follows full-amodal segmentation, annotators may
need to complete boundaries behind neighboring plants or field objects
even when the image provides little evidence for the hidden shape
~\cite{zhu2017semanticamodal,qi2019kins,zhan2024amodal}. EgoMaize
therefore defines an evidence-closed annotation workflow for this
setting: visible maize pixels are labeled by plant ownership, unreliable
visible regions go to ignore, and invisible regions are completed only
for locally constrained gaps where visible structures support the same
plant ownership and continuity. External occlusion and unverifiable hidden
regions are not completed.

\begin{figure}[!t]
\centering
\setlength{\tabcolsep}{2pt}
\begin{tabular}{ccc}
\includegraphics[width=0.31\linewidth]{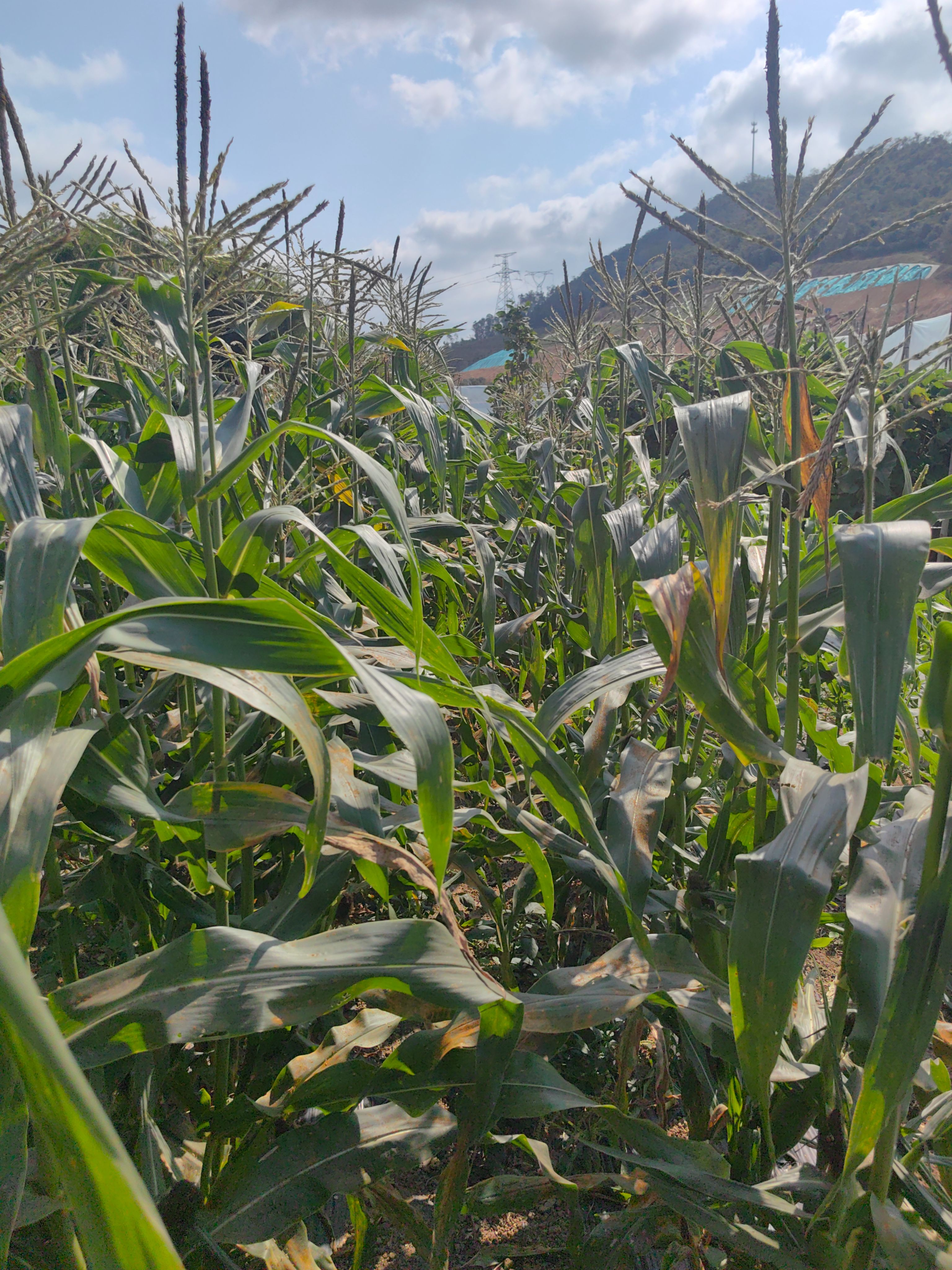} &
\includegraphics[width=0.31\linewidth]{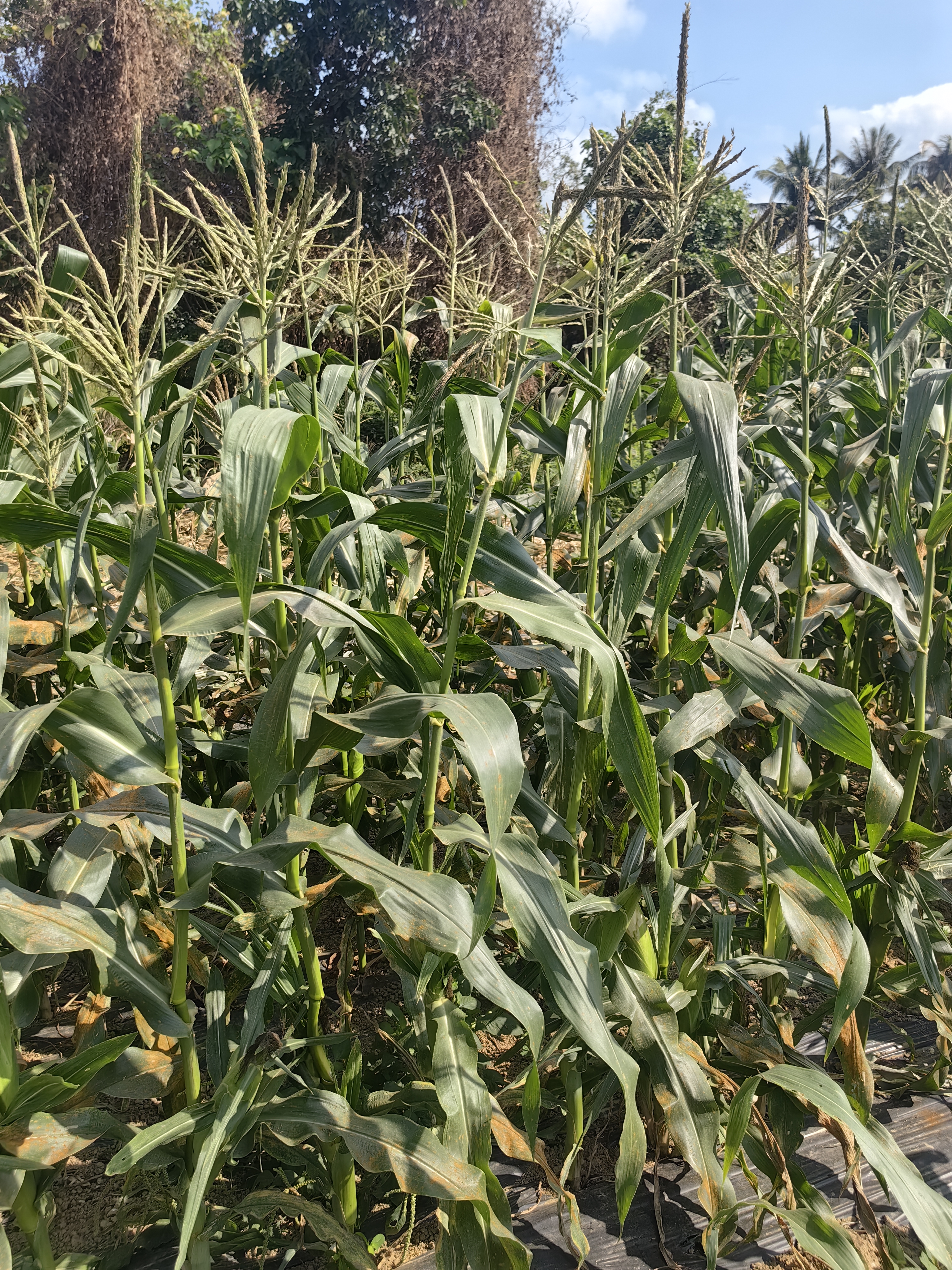} &
\includegraphics[width=0.31\linewidth]{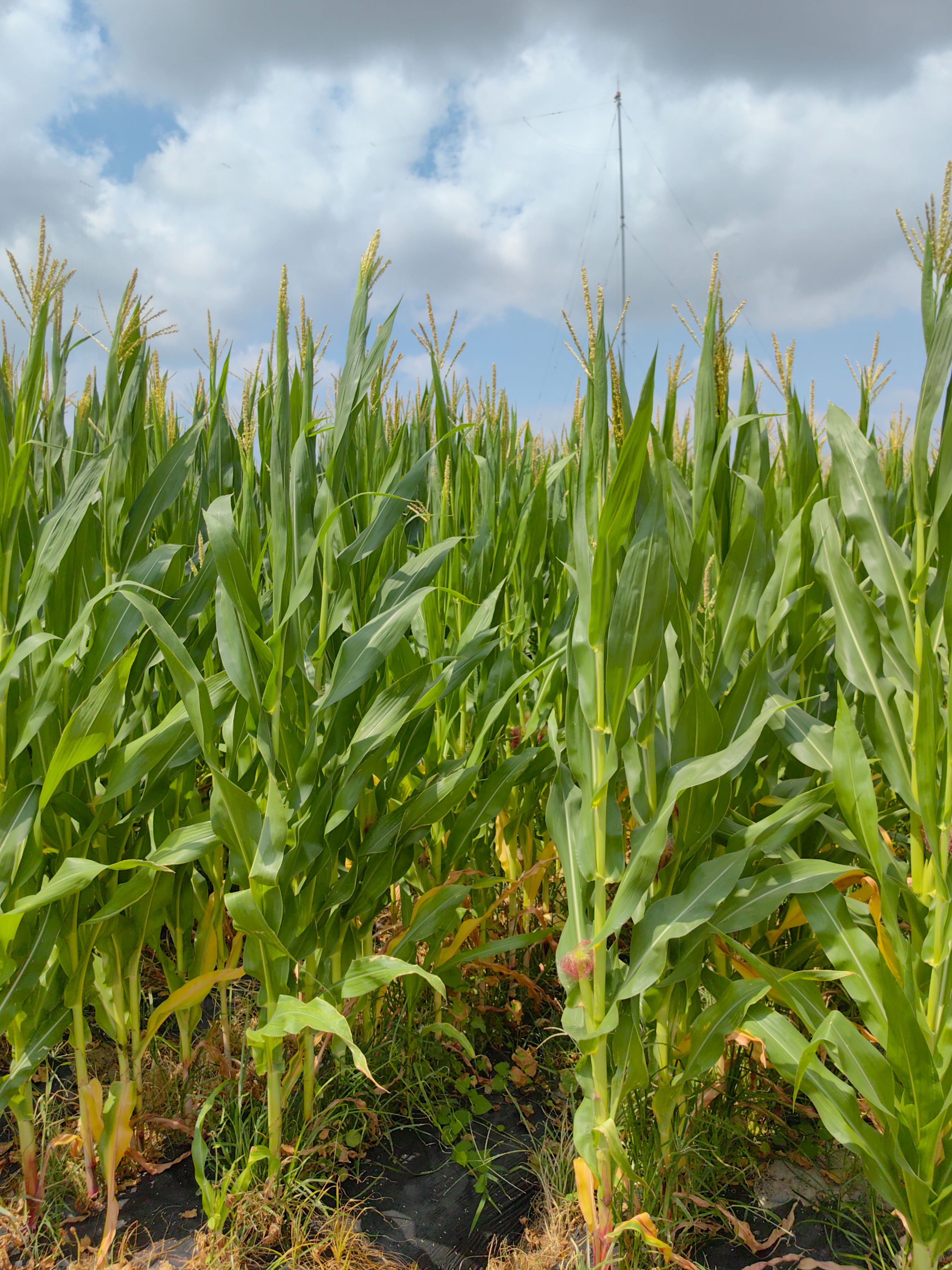} \\
(a) In-row breeding field & (b) Ridge-side breeding field & (c) Ridge-side seed production field \\
\end{tabular}
\caption{
Representative first-person post-seedling maize images.
}
\label{fig:intro_annotation}
\end{figure}

Figure~\ref{fig:intro_annotation} shows representative first-person
images in EgoMaize. These examples illustrate why the task is not simply
to segment maize foreground. A single physical plant may appear as
disconnected visible fragments, while adjacent visible regions may belong
to different plants. The core challenge is therefore ownership-consistent
instance delineation under dense same-class occlusion.

To make the benchmark diagnostic, we evaluate representative instance
segmentation baselines, analyze performance across occlusion levels, and
validate the annotation protocol through controlled artificial-occlusion
and inter-annotator consistency experiments. The baseline results show
that approximate localization and loose instance matching are easier than
high-quality ownership-consistent masks, especially as visibility becomes
limited. The annotation analyses further show that the evidence-closed
protocol is more consistent than unconstrained full-amodal completion in
this field setting. This leads to three contributions.

\noindent\textbf{Contributions.}
\begin{itemize}
\item We present EgoMaize, a compact egocentric benchmark for
post-seedling maize instance segmentation, specifically designed to
address plant-instance ownership under severe real-world field occlusion.

\item We establish an evidence-driven, evidence-closed annotation workflow
tailored to this setting. The workflow preserves intra-plant continuity
supported by local visual evidence, while suppressing speculative
full-amodal completion behind external occluders where empirical evidence
is absent.

\item We provide dataset-level analysis, representative instance
segmentation baselines, controlled artificial-occlusion validation, and
inter-annotator consistency experiments to characterize the difficulty of
ownership-consistent maize instance segmentation.
\end{itemize}

\section{Related Work}
\label{sec:related}

\subsection{Agricultural Vision Datasets and Field Phenotyping}

Agricultural vision datasets have expanded rapidly across crop--weed
segmentation, field phenotyping, and plant-level perception. PhenoBench
provides dense agricultural annotations and benchmarks for semantic
segmentation, plant panoptic segmentation, leaf instance segmentation,
and plant/leaf detection~\cite{weyler2024phenobench}. CropAndWeed
supports multimodal crop and weed perception for precision-agriculture
manipulation~\cite{steininger2023cropandweed}. WeedsGalore extends
maize-field crop/weed segmentation with multispectral and multitemporal
UAV imagery, while MFWD provides curated weed annotations in maize and
sorghum fields~\cite{celikkan2025weedsgalore,genze2024mfwd}. UAV-based
maize seedling instance segmentation further shows the value of
plant-level perception for early-stage crop monitoring
~\cite{geng2024maizeseedling}. In parallel, 3D plant resources such as
Pheno4D and Crops3D provide valuable geometric data for plant phenotyping
and agricultural perception~\cite{schunck2021pheno4d,Zhu2024Crops3D}.

These datasets are important, but most of them focus on overhead views,
crop--weed separation, seedling-stage plants, weed diversity, or 3D
geometry. They do not directly isolate first-person post-seedling maize
rows where many same-class organs overlap and must be assigned to the
correct plant instance. This distinction is relevant to agricultural
engineering. Ground robots and in-canopy sensing platforms are being used
to collect traits that are difficult to obtain from above-canopy sensing,
including ear height, stem diameter, and other plant-architecture
measurements~\cite{xu2022groundrobotreview,debruin2025maizeRobot}.
EgoMaize targets the 2D perception layer needed by such close-range
systems: ownership-aware maize plant instance segmentation under severe
field occlusion.

\subsection{Annotation under Occlusion}

Many segmentation benchmarks rely on modal masks, where only visible
object pixels are labeled. Modal labels are reproducible, but they can
fragment one physical object into multiple disconnected visible parts.
Amodal segmentation instead labels the full object extent behind
occluders~\cite{zhu2017semanticamodal,qi2019kins}. This can encourage
object-level completion, but manually inferred amodal ground truth in
real images becomes subjective when hidden boundaries are not constrained
by evidence. Recent work has emphasized this issue and used 3D information
to obtain more objective amodal targets~\cite{zhan2024amodal}.

EgoMaize is related to this line of work, but it is not intended as a
general amodal segmentation dataset. Its difficulty comes from a
single-scene agricultural setting with many same-class, elongated plant
instances. Under this setting, the key question is not complete hidden
shape recovery, but whether visible and locally recoverable maize regions
can be reliably assigned to the same plant instance.

Ignore or void regions provide another way to avoid forcing uncertain
pixels into foreground or background. COCO and Cityscapes use crowd, void,
or ignore labels to reduce penalties in ambiguous regions
~\cite{lin2014coco,cordts2016cityscapes}. In EgoMaize, ignore is part of
the label schema rather than only a cleanup label: recognizable maize
regions with uncertain ownership or unreliable boundaries are excluded
from both positive instances and background.

Occlusion-aware agricultural perception has also been studied for compact
fruits. Amodal tomato segmentation aims to recover complete fruit shape
under occlusion~\cite{yang2024tomatoamodal,li2025cgaasnet}. Post-seedling
maize differs in geometry and annotation objective. A tomato is a compact
object whose hidden contour may be a natural target; a maize plant is an
elongated structure composed of interleaved organs, and the central
problem is assigning regions to the correct plant instance without
inventing unsupported cross-plant boundaries.

\section{EgoMaize Dataset}
\label{sec:dataset}

This section reports how EgoMaize is defined and instantiated as a
dataset. We first position the dataset relative to existing agricultural
datasets and define the target task and its challenges. We then describe
the dataset scope, annotation schema and workflow, split, and statistics.

\subsection{Task, Challenges, and Dataset Positioning}
\label{sec:dataset_comparison}

EgoMaize defines a plant-level instance segmentation task for first-person
post-seedling maize images. Given a close-range field image, the model is
expected to predict ownership-consistent maize plant instance masks. The
dataset also provides plant-owned stem and tassel cues, which are not
independent object instances but auxiliary fields linked to their parent
plant. Regions that are recognizable as maize but have unreliable
ownership or boundaries are marked as ignore and excluded from training
and evaluation.

The main challenge is ownership-consistent mask delineation under severe
same-class overlap. Unlike crop--weed segmentation or overhead seedling
detection, EgoMaize contains many visually similar maize plants in the
same image. Their stems, leaves, and tassels are elongated, repetitive,
and frequently interleaved. A single physical plant may appear as
disconnected visible fragments, while adjacent visible regions may belong
to different plants. Therefore, the task is not merely to segment maize
foreground or count plant centers, but to assign visible and locally
recoverable regions to the correct plant instance.

EgoMaize is designed as an agricultural instance segmentation benchmark
rather than a general amodal segmentation dataset. Its main gap is the
combination of first-person field views, post-seedling maize plants,
same-class plant-instance ownership, and severe field occlusion. This
combination is important for close-range phenotyping and robotic crop
monitoring, where the system must associate visible stems, tassels, and
fragmented plant regions with the correct plant instance. We therefore
compare EgoMaize primarily with agricultural vision datasets. The
comparison focuses on the scene, instance unit, ownership ambiguity, and
annotation fields, rather than dataset scale.

Table~\ref{tab:agri_datasets} shows that existing agricultural datasets
mainly address crop--weed recognition, UAV-based plant monitoring,
seedling instances, leaf-level annotation, weed diversity, or 3D plant
geometry. These settings are valuable, but they do not directly isolate
the setting of close-range post-seedling maize rows where many visually
similar organs overlap and plant-instance ownership must be resolved from
2D first-person evidence.

\begin{table}[t]
\centering
\scriptsize
\setlength{\tabcolsep}{2.0pt}
\renewcommand{\arraystretch}{1.12}
\caption{
Comparison with representative agricultural vision datasets.
}
\label{tab:agri_datasets}
\begin{tabularx}{\linewidth}{
@{}l
L{0.14\linewidth}
L{0.14\linewidth}
L{0.15\linewidth}
L{0.18\linewidth}
X@{}}
\toprule
Dataset
& View / modality
& Stage / target
& Instance unit
& Same-class overlap / ownership
& Annotation fields \\
\midrule

PhenoBench~\cite{weyler2024phenobench}
& UAV / field
& crops and weeds
& plant / leaf
& mainly top-down; less first-person plant ownership ambiguity
& semantic, panoptic, plant / leaf labels \\

CropAndWeed~\cite{steininger2023cropandweed}
& field / robotic
& crops and weeds
& crop / weed regions
& focuses on crop--weed separation rather than same-crop plant ownership
& multimodal crop--weed labels \\

WeedsGalore~\cite{celikkan2025weedsgalore}
& UAV / multispectral
& maize-field weeds
& crop--weed regions
& crop and weed perception, not maize plant-instance ownership
& crop--weed segmentation \\

MFWD~\cite{genze2024mfwd}
& field images
& maize / sorghum weeds
& weed regions
& focuses on weed diversity rather than post-seedling maize instances
& curated weed annotations \\

UAV maize seedling~\cite{geng2024maizeseedling}
& UAV / overhead
& maize seedlings
& seedling plant
& overhead early-stage plants; less organ-level overlap
& seedling plant instances \\

Pheno4D~\cite{schunck2021pheno4d}
& 3D / 4D scan
& plants
& plant / organ geometry
& ownership can be supported by 3D geometry, not 2D first-person evidence
& plant and organ-level 3D structure \\

Crops3D~\cite{Zhu2024Crops3D}
& 3D scan
& crops
& 3D crop objects
& different modality and acquisition cost
& 3D crop perception labels \\

EgoMaize
& first-person RGB
& post-seedling maize
& maize plant
& severe same-class overlap and plant-instance ownership ambiguity
& plant masks, stem/tassel cues, ignore regions \\
\bottomrule
\end{tabularx}
\end{table}

The comparison shows that EgoMaize is complementary to existing
agricultural datasets. UAV datasets are effective for crop counting,
crop--weed separation, and canopy-level monitoring, but overhead views
reduce the same-class ownership ambiguity observed from within maize rows.
3D datasets provide stronger geometry but require different acquisition
hardware and do not define a 2D first-person instance-mask benchmark.
EgoMaize targets the missing 2D setting where post-seedling maize organs
from multiple plants overlap in the camera view, and annotation must
separate visible regions, evidence-closed local gaps, and unreliable
maize pixels.

\subsection{Dataset Scope}
\label{sec:dataset_scope}

EgoMaize contains first-person, close-range images of post-seedling maize
plants captured from field-level viewpoints. The images include row-level
and oblique perspectives from within the field rather than only side-view
or overhead views. This setting exposes dense overlap among stems, leaves,
tassels, and neighboring plant instances, making it suitable for studying
ownership-aware plant instance segmentation under severe occlusion.

The current release contains 301 RGB images. The original image resolution
is $3072\times4096$. All images are annotated under the workflow defined
in Section~\ref{sec:annotation_workflow}. The released annotations are
centered on maize plant instances, plant-owned stem and tassel fields, and
image-level ignore regions. EgoMaize is therefore intended as a focused
pretrained fine-tuning and evaluation benchmark for a specialized target
domain, rather than a large-scale pretraining corpus.

\begin{figure}[!t]
\centering
\setlength{\tabcolsep}{2pt}
\begin{tabular}{cccc}
\includegraphics[height=0.2\textheight,keepaspectratio]{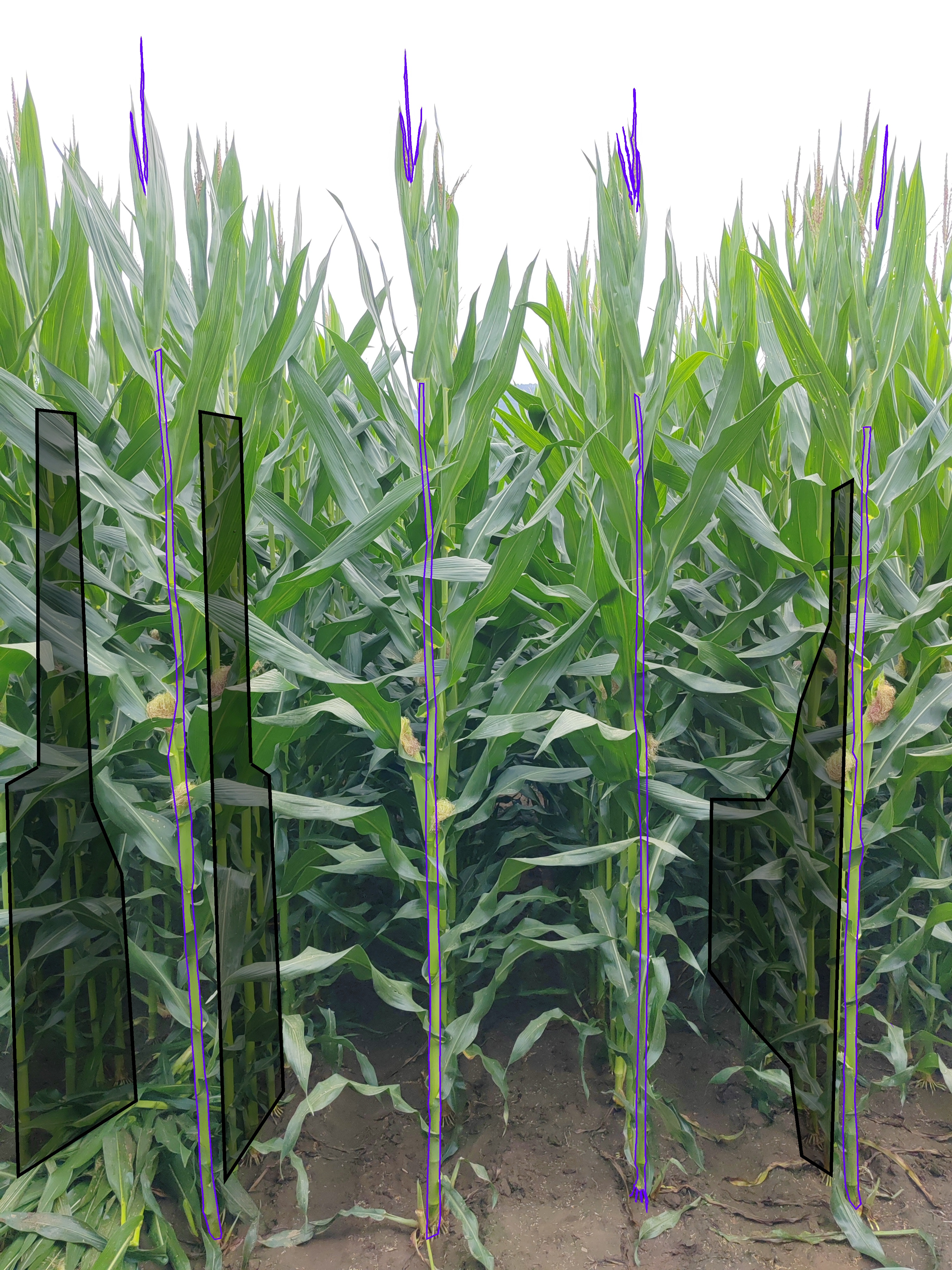} &
\includegraphics[height=0.2\textheight,keepaspectratio]{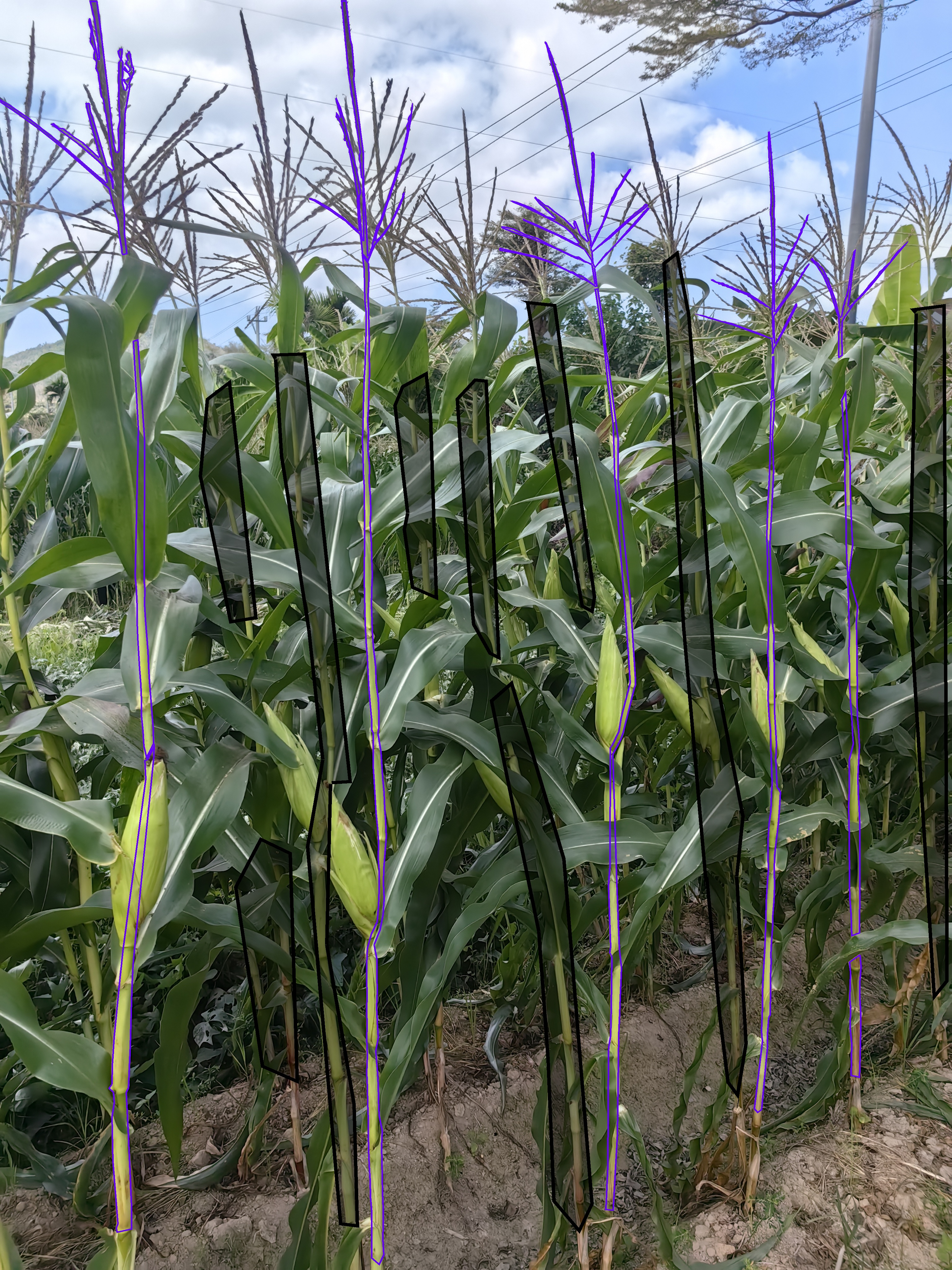} &
\includegraphics[height=0.2\textheight,keepaspectratio]{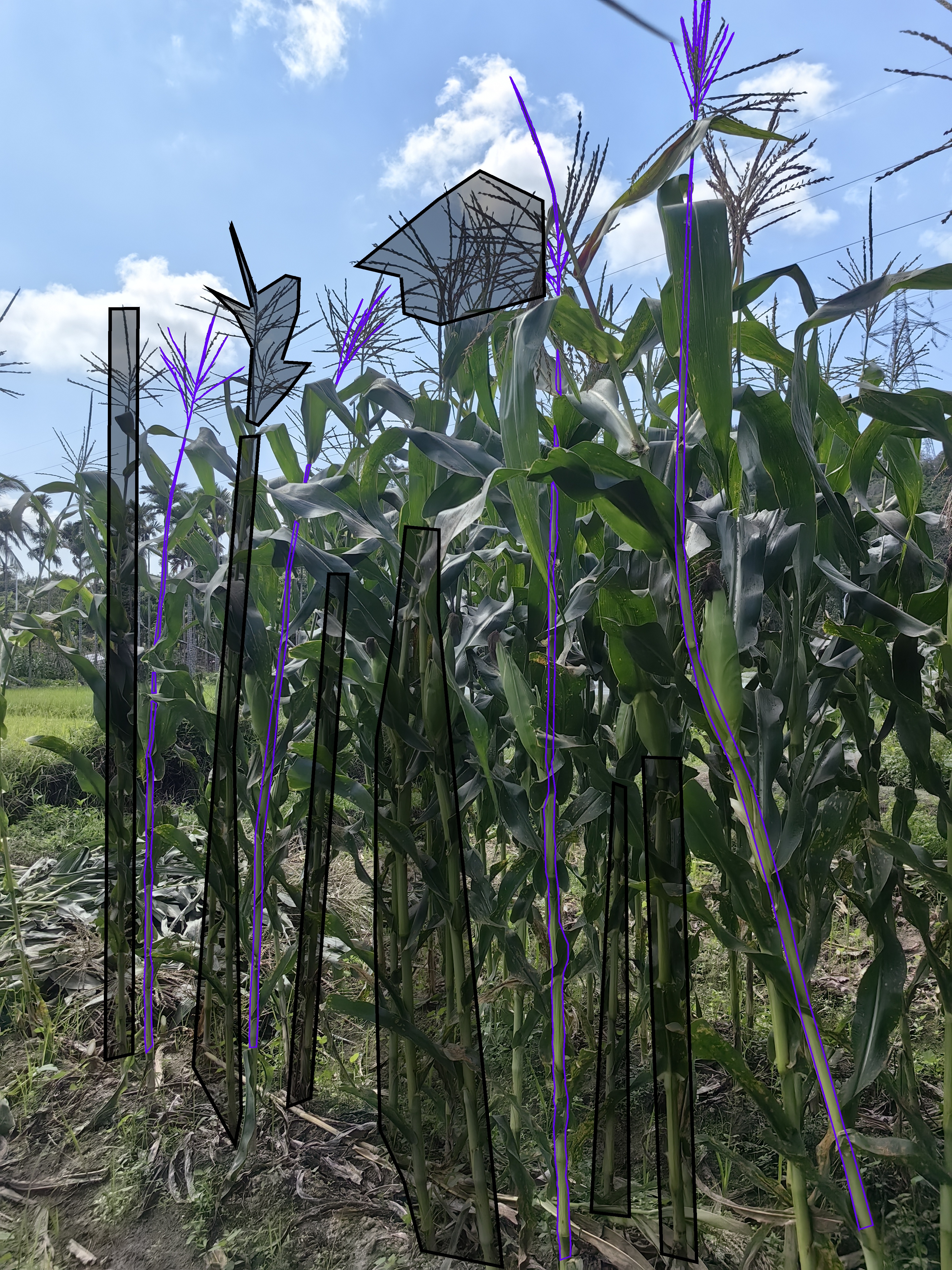} &
\includegraphics[height=0.2\textheight,keepaspectratio]{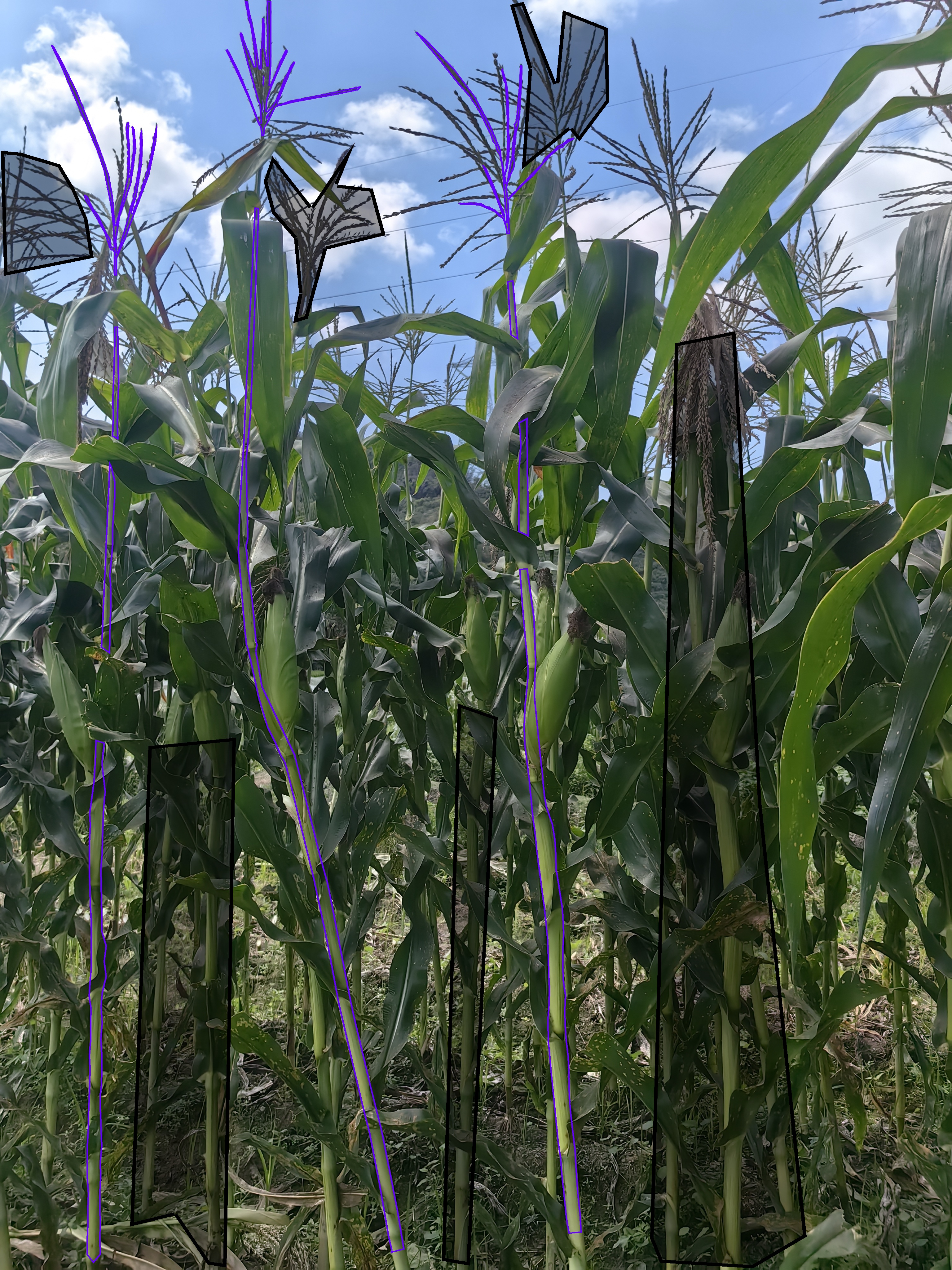}
\end{tabular}
\caption{
Representative EgoMaize annotation samples.
}
\label{fig:annotation_samples}
\end{figure}

Figure~\ref{fig:annotation_samples} shows representative annotation
samples from EgoMaize. These samples illustrate the released annotation
fields and the main visual difficulty of the dataset: multiple
post-seedling maize plants from the same class overlap heavily, while stem
and tassel cues must be associated with the correct plant instance.
Ownership-ambiguous maize regions are marked as ignore rather than forced
into either plant masks or background.

\subsection{Annotation Schema and Workflow}
\label{sec:annotation_workflow}

Each plant instance is represented as $y_i=(M_i^{+},\mathcal{P}_i)$,
where $M_i^{+}$ is the positive plant-instance mask and
$\mathcal{P}_i=\{M_i^{stem},M_i^{tas}\}$ stores plant-owned stem and
tassel cues. These organ cues are linked to the parent plant rather than
treated as independent object instances. The dataset also provides an
image-level ignore mask $\Omega^{ign}$ for recognizable maize pixels with
uncertain ownership, unreliable boundaries, or severe occlusion.

The annotation workflow is designed to adapt existing modal, amodal, and
ignore-label conventions to this crop setting.
We refer to the resulting occlusion-handling rule as the evidence-closed annotation protocol. Visible maize regions are
annotated according to reliable plant ownership; visible regions with
uncertain ownership or unreliable boundaries are assigned to ignore rather
than background. Invisible regions are completed only when they form
evidence-closed local gaps.
Evidence closure is judged from visible image evidence, the actual
occlusion configuration, and domain knowledge of maize structure. In
practice, short self-occlusions can often be evidence-closed because
maize shoot architecture follows regular organ-level organization and can
be represented by 3D architectural or phytomer-based models
~\cite{fournier1998maize3d,wen2021phytomer}. Visible stem segments and
nearby leaf-sheath or leaf-attachment cues can therefore support
same-plant ownership and local continuity.
By contrast, large hidden areas, long-range interruptions, external
occlusion, or cases where
relative depth, scale, or image position makes the stem structure
unverifiable are not completed. Such regions are left outside positive
instances, and recognizable but unreliable maize pixels are marked as
ignore.

\begin{figure}[!t]
\centering
\includegraphics[width=0.9\linewidth,height=0.9\textheight,keepaspectratio]{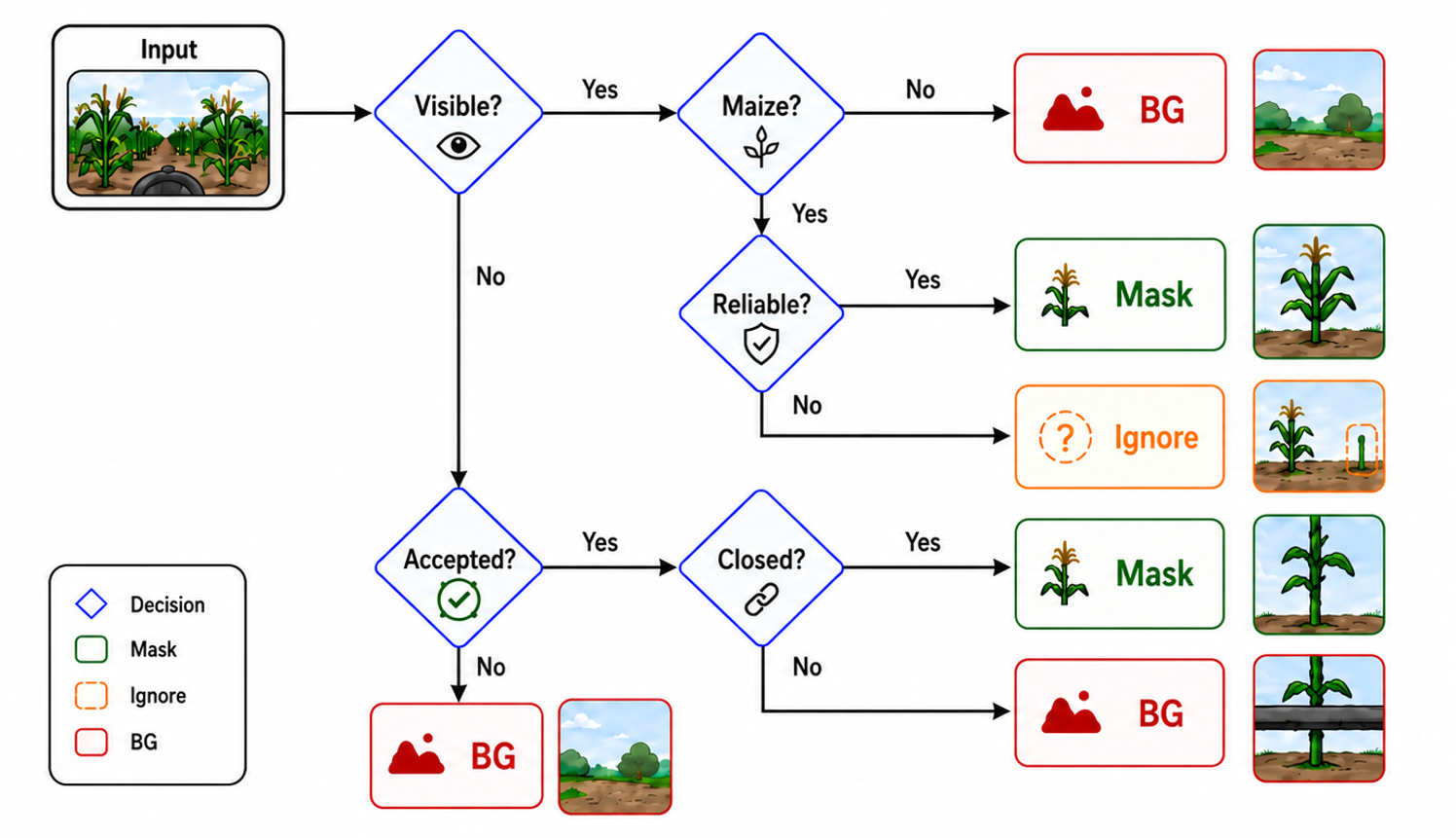}
\caption{
Annotation decision flow in EgoMaize. Trained annotators perform the
primary annotation, while domain experts and reviewers support protocol
clarification and quality control. The workflow keeps annotation focused
on plant-instance ownership rather than visible connected components
alone.
}
\label{fig:annotation_flow}
\end{figure}

Figure~\ref{fig:annotation_flow} summarizes this decision workflow.

Annotation is performed by four trained annotators who were instructed on
the EgoMaize annotation workflow before labeling. The training stage
focused on plant-instance ownership, stem/tassel assignment,
evidence-closed completion, external occlusion handling, and ignore-region
use. Ambiguous cases were discussed and converted into shared annotation
rules before full annotation.

Annotation is performed in a human-in-the-loop manner. Interactive
segmentation tools, including SAM~\cite{kirillov2023sam}, may be used to
generate initial mask proposals and reduce manual drawing effort. These
proposals are immediately followed by manual refinement, where annotators
correct plant-instance ownership, mask boundaries, stem/tassel assignment,
evidence-closed completion, and ignore regions according to the workflow.
The released annotations are therefore not raw SAM outputs. SAM is used as
an annotation aid, not as an automatic label generator or an additional
supervision source.
We also release Maize PreSeg Tool, the annotation and pre-segmentation
platform used in this work, at
\url{https://github.com/JaaaaaaaD/Maize_PreSeg_Tool}. The tool supports
manual annotation, SAM-assisted pre-annotation, COCO-format import/export,
instance-level refinement, and correction-trace export.

Due to dense overlap, organ ownership ambiguity, and the need to refine
plant masks, stem/tassel cues, and ignore regions jointly, EgoMaize
annotation is time-consuming. In our annotation process, a single image
takes approximately 25 minutes to complete on average. This cost is a main
reason why EgoMaize is designed as a compact but densely annotated
benchmark rather than a large-scale weakly annotated dataset.

Quality control focuses on three checks: whether plant-instance ownership
is consistent, whether stem and tassel fields are linked to the correct
plant instance, and whether uncertain maize regions are assigned to ignore
rather than forced into foreground or background. Disagreement and
difficult cases are reviewed against the shared workflow before release.
This review process keeps the dataset aligned with the plant-instance-level
supervision target.

\subsection{Dataset Split and Statistics}
\label{sec:dataset_split}
\label{sec:dataset_statistics}

We provide a fixed train/validation/test split for reproducible
benchmarking. Since EgoMaize is designed around occlusion and
plant-instance ownership, the split is not produced by random image
sampling alone. Instead, we use a score-stratified procedure so that
images with different instance density, occlusion difficulty, and ignore
coverage are represented across the three subsets.

For each image, we compute a compact difficulty score. Let $c$ be the
number of plant instances in an image, and let
$n=(c-c_{\min})/(c_{\max}-c_{\min})$ be its normalized instance count. The
sampling score is
\begin{equation}
s=(1+0.5n)\,2(\bar{o}+o_{\max})+u,
\end{equation}
where $\bar{o}$ and $o_{\max}$ denote the mean and maximum height-wise
occlusion ratios of plant instances in the image, and $u$ denotes the
ignore-region union area ratio. Images are grouped by this score and then
assigned to train, validation, and test subsets in a stratified manner.
This prevents the validation or test set from being dominated by easy
low-occlusion images.

\begin{figure}[!t]
\centering
\begin{minipage}{0.48\linewidth}
\centering
\includegraphics[height=0.25\textheight,keepaspectratio]{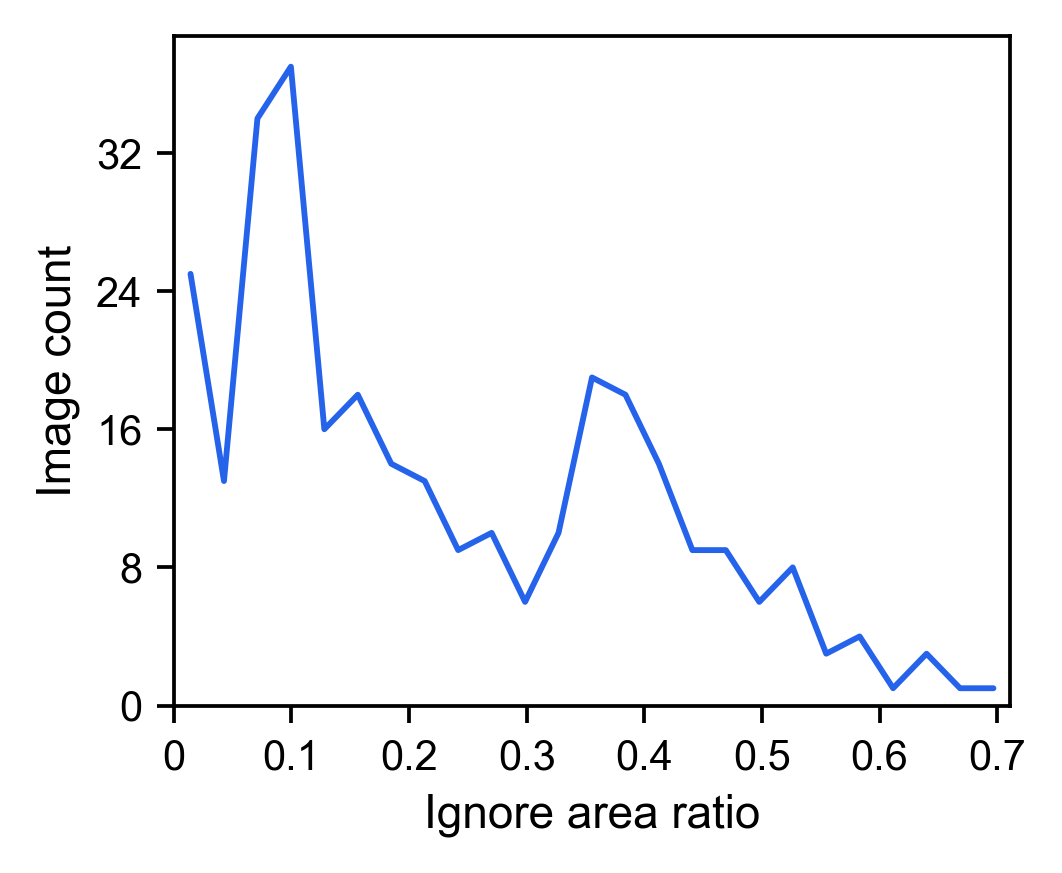}
\end{minipage}
\hfill
\begin{minipage}{0.48\linewidth}
\centering
\includegraphics[height=0.25\textheight,keepaspectratio]{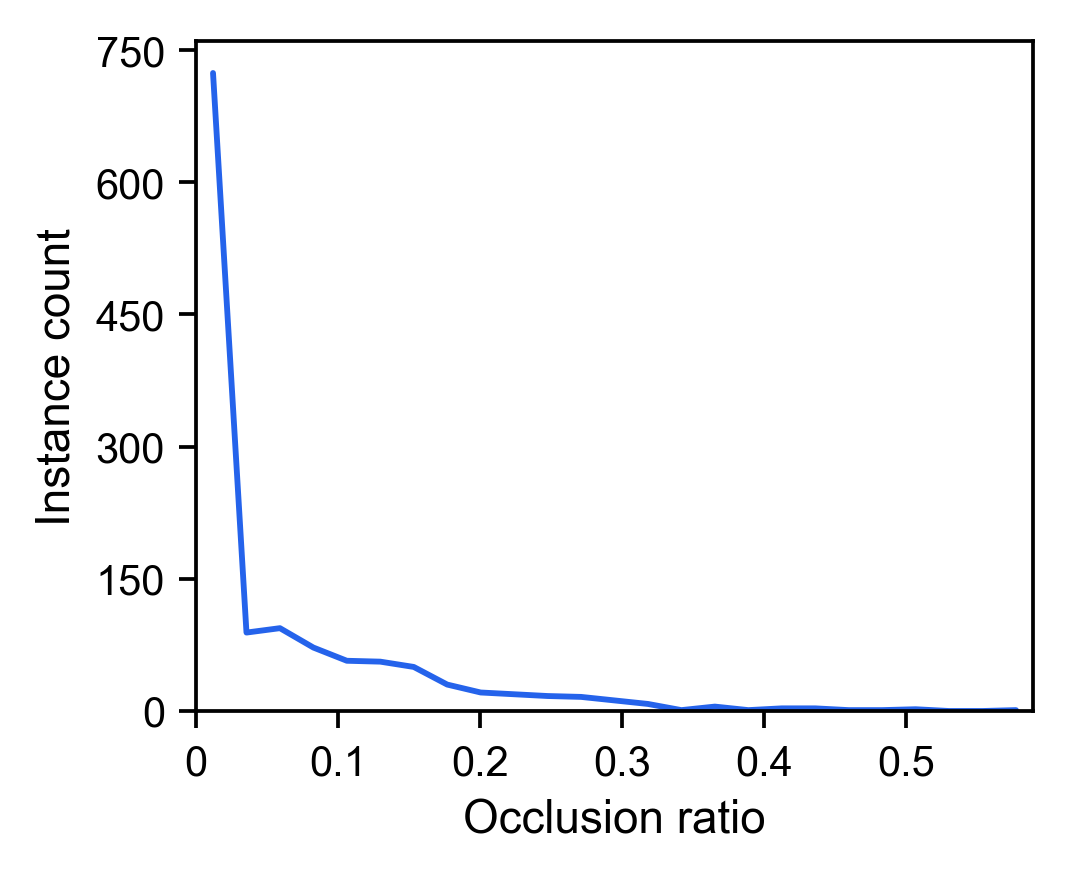}
\end{minipage}
\vspace{-1mm}
\caption{
Dataset difficulty distributions.
}
\label{fig:dataset_difficulty}
\vspace{-2mm}
\end{figure}

Table~\ref{tab:split} reports the resulting fixed split used in the
benchmark experiments. Table~\ref{tab:stats} further reports the dataset
statistics, and Figure~\ref{fig:dataset_difficulty} shows the
image-level ignore-region distribution and instance-level height-wise
occlusion distribution. Organ fields are linked to plant instances rather
than treated as independent object instances. These statistics indicate
that uncertain regions are common and that occlusion difficulty follows a
long-tailed pattern.

\begin{table}[t]
\centering
\small
\caption{
Fixed dataset split.
}
\label{tab:split}
\begin{tabular}{lcc}
\toprule
Split & Images & Positive instances \\
\midrule
Train & 211 & 906 \\
Validation & 45 & 200 \\
Test & 45 & 170 \\
\midrule
Total & 301 & 1,276 \\
\bottomrule
\end{tabular}
\end{table}

\begin{table}[t]
\centering
\small
\caption{
Dataset statistics.
}
\label{tab:stats}
\begin{tabular}{lc}
\toprule
Statistic & Value \\
\midrule
Images & 301 \\
Original image resolution & $3072\times4096$ \\
Plant instances & 1,276 \\
Organ-field annotations & 9,043 \\
Average plant instances per image & 4.24 \\
Average organ-field annotations per plant & 7.09 \\
Ignore-region annotations & 2,731 \\
Images with ignore regions & 282 \\
Mean ignore union ratio per image & 0.2311 \\
Median ignore union ratio per image & 0.1843 \\
Mean plant-instance area (pixels) & 96,555.80 \\
Mean plant-instance area ratio & 0.0078 \\
\bottomrule
\end{tabular}
\end{table}

\section{Benchmark and Annotation Analysis}
\label{sec:experiments}

We evaluate EgoMaize as a benchmark for egocentric post-seedling maize
instance segmentation. The main experiment is a full-dataset benchmark
with representative instance segmentation models. We then provide an
occlusion-level breakdown, an ignore-region sensitivity analysis, a
controlled artificial-occlusion validation, and an annotation
reproducibility analysis through inter-annotator consistency. The goal is
not to propose a new segmentation architecture, but to characterize what
current models can and cannot learn from this dataset.

\subsection{Evaluation Setup}
\label{sec:eval_setup}

All models are evaluated on the fixed split in
Section~\ref{sec:dataset_split} with the same ignore-mask handling.
Ignore pixels are excluded from both training losses and metric
calculation. We report five metrics: mIoU, AP$_{25}$, AP$_{50}$, count
MAE, and bounding-box center distance. mIoU measures coarse
foreground-region overlap. AP$_{25}$ and AP$_{50}$ measure instance-level
mask matching under loose and stricter IoU thresholds. Count MAE measures
plant counting error after thresholding. Bounding-box center distance is
reported in pixels and measures the average Euclidean distance between
the center of each ground-truth instance bounding box and the center of
its nearest predicted instance bounding box.

We benchmark representative baselines covering one-stage, two-stage,
boundary-refinement, occlusion-aware, query-based, and crop-refinement
segmentation: YOLOv8m-seg~\cite{jocher2023ultralytics},
YOLO11m-seg~\cite{ultralytics_yolo11},
Mask R-CNN~\cite{he2017maskrcnn}, PointRend~\cite{kirillov2020pointrend},
BCNet~\cite{ke2021bcnet}, Mask2Former R50 and Mask2Former
Swin-B~\cite{cheng2022mask2former}, CropFormer~\cite{qi2023highqualityentity},
and DI-MaskDINO-R50~\cite{nan2024dimaskdino}. This model coverage is used
to expose different failure modes rather than to select a single best
architecture.

All baselines use their recommended pretrained initialization and intended
effective configurations whenever available. We follow the official
training configuration of each method, including optimizer, learning-rate
schedule, input preprocessing, and data augmentation settings, and only
adapt dataset paths, class definitions, and ignore-mask handling to
EgoMaize. For Mask2Former Swin-B, we use the official COCO-pretrained
checkpoint and released configuration. Detailed initialization sources,
input sizes, training lengths, and model-selection rules are reported in
the supplementary material.

\subsection{Overall Baseline Benchmark}
\label{sec:overall_baseline}

Table~\ref{tab:full_baseline} reports the overall benchmark results on
EgoMaize. The table compares representative models under the same split,
annotation, ignore policy, and metric implementation.

\begin{table}[t]
\centering
\small
\setlength{\tabcolsep}{2.5pt}
\renewcommand{\arraystretch}{1.12}
\caption{
Overall baseline benchmark.
}
\label{tab:full_baseline}
\begin{tabular}{@{}l l c c c c c@{}}
\toprule
\rowcolor{HeaderBlue}
Model & Type
& mIoU$\uparrow$
& AP$_{25}\uparrow$
& AP$_{50}\uparrow$
& C-MAE$\downarrow$
& C-dist.$\downarrow$ \\
\midrule

YOLOv8m-seg & one-stage
& 0.0633 & 0.1146 & 0.0000 & 15.60
& \cellcolor{SecondYellow}\underline{101.95} \\

YOLO11m-seg & one-stage+
& 0.0649 & 0.1124 & 0.0000 & 14.84
& \cellcolor{BestGreen}\textbf{95.77} \\

Mask R-CNN & two-stage
& 0.2575 & 0.3688 & 0.0053 & 4.27
& 125.22 \\

PointRend & boundary
& 0.3544
& 0.6307
& 0.1291
& 8.93 & 246.50 \\

BCNet & occlusion
& 0.2635 & 0.4818 & 0.0185 & 2.24 & 299.38 \\

Mask2Former R50 & query
& 0.2493
& \cellcolor{SecondYellow}\underline{0.6707}
& 0.2656
& \cellcolor{SecondYellow}\underline{1.93}
& 168.59 \\

Mask2Former Swin-B & query+
& \cellcolor{BestGreen}\textbf{0.5014}
& \cellcolor{BestGreen}\textbf{0.8024}
& \cellcolor{SecondYellow}\underline{0.5586}
& 2.18
& 188.94 \\

CropFormer & crop-refine
& \cellcolor{SecondYellow}\underline{0.4061}
& 0.6586
& \cellcolor{BestGreen}\textbf{0.6585}
& \cellcolor{BestGreen}\textbf{1.40}
& 140.15 \\

DI-MaskDINO-R50 & query++
& 0.0920 & 0.0430 & 0.0000 & 23.17 & 849.76 \\
\bottomrule
\end{tabular}
\end{table}

\paragraph{Model behavior.}
Table~\ref{tab:full_baseline} shows that different baselines perform well
on different metrics, which indicates that EgoMaize is not captured by a
single notion of segmentation quality. Mask2Former Swin-B obtains the
highest mIoU and AP$_{25}$, showing that a stronger pretrained query-based
model can substantially improve coarse foreground overlap and loose
instance matching. CropFormer obtains the best AP$_{50}$ and count MAE,
indicating that high-resolution crop refinement remains effective for
stricter mask agreement and plant counting. YOLO11m-seg obtains the
smallest center distance, but both YOLO variants have very low mask AP,
suggesting that center localization is easier than ownership-consistent
mask delineation.

The gap between AP$_{25}$ and AP$_{50}$ remains substantial for most
models. For the corrected Mask2Former Swin-B, AP$_{25}$ reaches 0.8024,
while AP$_{50}$ is 0.5586. This indicates that the model can often recover
coarse plant instances, but stricter ownership-consistent mask agreement
remains difficult. This is consistent with the visual properties of
EgoMaize: maize plants are thin, fragmented, repetitive, and often
interleaved with neighboring plants.

Overall, the benchmark shows that current instance segmentation models can
partly localize maize plants and sometimes recover coarse instances, but
ownership-consistent mask prediction remains challenging. The added recent
baselines do not change this conclusion: YOLO11m-seg does not improve
strict mask AP, and DI-MaskDINO-R50 shows severe over-prediction under the
current configuration. DI-MaskDINO-R50 is reported as a three-seed
diagnostic mean under the same EgoMaize evaluator; full per-seed results
are provided in the supplementary material. These results motivate
EgoMaize as a diagnostic benchmark for fine structures, same-class
overlap, occlusion, and ignore-region handling, rather than as a setting
already solved by existing architectures.

\subsection{Occlusion-Level Breakdown}
\label{sec:occlusion_breakdown}

Overall scores hide where models fail. Table~\ref{tab:occlusion_breakdown_ap25}
reports AP$_{25}$ under different height-wise occlusion levels. Each
ground-truth instance is assigned to one occlusion bin according to the
height-wise occlusion ratio defined in Section~\ref{sec:dataset_statistics}.
The $r>0.5$ bin is omitted because valid plant instances in EgoMaize are
kept below this threshold by the annotation workflow.

\begin{table}[t]
\centering
\small
\setlength{\tabcolsep}{4.2pt}
\renewcommand{\arraystretch}{1.12}
\caption{
AP$_{25}$ breakdown by height-wise occlusion ratio.
}
\label{tab:occlusion_breakdown_ap25}
\begin{tabular}{@{}l c c c c@{}}
\toprule
\rowcolor{HeaderBlue}
Model
& $0$
& $(0,0.1]$
& $(0.1,0.3]$
& $(0.3,0.5]$ \\
\midrule

YOLOv8m-seg
& 0.0008 & 0.1165 & 0.0002 & 0.0000 \\

Mask R-CNN
& 0.2751 & 0.3587 & 0.0875 & 0.0254 \\

PointRend
& 0.6129
& 0.5310
& 0.4399
& \cellcolor{BestGreen}\textbf{0.4567} \\

BCNet
& 0.4885 & 0.4666 & 0.2255 & 0.1047 \\

Mask2Former R50
& \cellcolor{SecondYellow}\underline{0.6403}
& \cellcolor{SecondYellow}\underline{0.6748}
& \cellcolor{SecondYellow}\underline{0.5655}
& \cellcolor{SecondYellow}\underline{0.4196} \\

Mask2Former Swin-B
& \cellcolor{BestGreen}\textbf{0.7171}
& \cellcolor{BestGreen}\textbf{0.7171}
& 0.4133
& 0.0915 \\

CropFormer
& 0.6373
& 0.6737
& \cellcolor{BestGreen}\textbf{0.6262}
& 0.2573 \\
\bottomrule
\end{tabular}
\end{table}

\paragraph{Occlusion analysis.}
Table~\ref{tab:occlusion_breakdown_ap25} reports how AP$_{25}$ changes
across height-wise occlusion levels. The results show that occlusion is an
important source of difficulty in EgoMaize. The corrected Mask2Former
Swin-B performs strongly in the no-occlusion and light-occlusion bins, but
drops from 0.7171 at $r=0$ to 0.0915 in the $(0.3,0.5]$ bin. This shows
that the previous abnormal collapse is not the source of the occlusion
trend: even a strong pretrained query-based model degrades sharply when
visibility becomes limited.

The trend is not strictly monotonic for every architecture because each
bin still contains variation in plant size, shape, visible structure, and
neighboring-instance ambiguity. However, no model maintains uniformly high
AP$_{25}$ across all visibility levels. The severe-occlusion bin remains
difficult even under the loose AP$_{25}$ threshold, indicating that
fragmented visibility and same-class overlap continue to challenge
instance matching.

Overall, the occlusion-level breakdown complements the overall benchmark:
it shows that performance differences are not only caused by global model
quality, but also by how predictions behave under different visibility
conditions. This supports the use of EgoMaize as a diagnostic benchmark
for ownership-consistent segmentation under field occlusion.

\subsection{Ignore-Region Sensitivity}
\label{sec:ignore_sensitivity}

Ignore regions are used for recognizable maize pixels whose plant
ownership or boundary cannot be reliably verified. To examine whether the
ignore policy artificially inflates the evaluation, we conduct a strict
diagnostic evaluation on the corrected Mask2Former Swin-B predictions by
treating all ignore pixels as background.

\begin{table}[t]
\centering
\small
\caption{
Sensitivity to ignore-region treatment for corrected Mask2Former Swin-B.
}
\label{tab:ignore_sensitivity}
\begin{tabular}{lccc}
\toprule
Ignore policy & mIoU$\uparrow$ & AP$_{25}\uparrow$ & AP$_{50}\uparrow$ \\
\midrule
Remove ignore pixels & 0.5014 & 0.8024 & 0.5586 \\
Treat ignore as background & 0.4087 & 0.7885 & 0.5344 \\
\bottomrule
\end{tabular}
\end{table}

Table~\ref{tab:ignore_sensitivity} shows that AP$_{25}$ and AP$_{50}$ are
only mildly affected by the stricter policy, changing from
0.8024/0.5586 to 0.7885/0.5344. Although 22.76\% of the predicted
foreground area lies inside ignore regions, predictions concentrated in
ignore regions do not explain the reported AP: among the 76 predictions
with more than half of their area inside ignore, 75 have best valid-region
GT IoU below 0.25 and none reaches 0.50. Predictions expanding into valid
background are still counted as false positives. Therefore, ignore regions
avoid supervising unverifiable maize ownership rather than rewarding
coarse masks.

\subsection{Controlled Artificial-Occlusion Validation}
\label{sec:controlled_occlusion}

We further construct a small controlled validation set at the Xiaotangshan
maize breeding base to evaluate annotation behavior under external
occlusion. The set contains 10 maize plants from different cultivars and
with different visible traits. For each plant, we captured three views:
an exposed reference view, a local self-occlusion view, and an artificial
external-occlusion view. The external occlusion was manually introduced to
simulate occlusion by another plant or object. For the reference view,
artificial occluders were removed and self-occluding leaves were moved
away as much as possible, so that exposed stem and tassel regions could be
annotated as a controlled reference.

\begin{figure}[!t]
\centering
\setlength{\tabcolsep}{2pt}
\begin{tabular}{ccc}
\includegraphics[height=0.25\textheight,keepaspectratio]{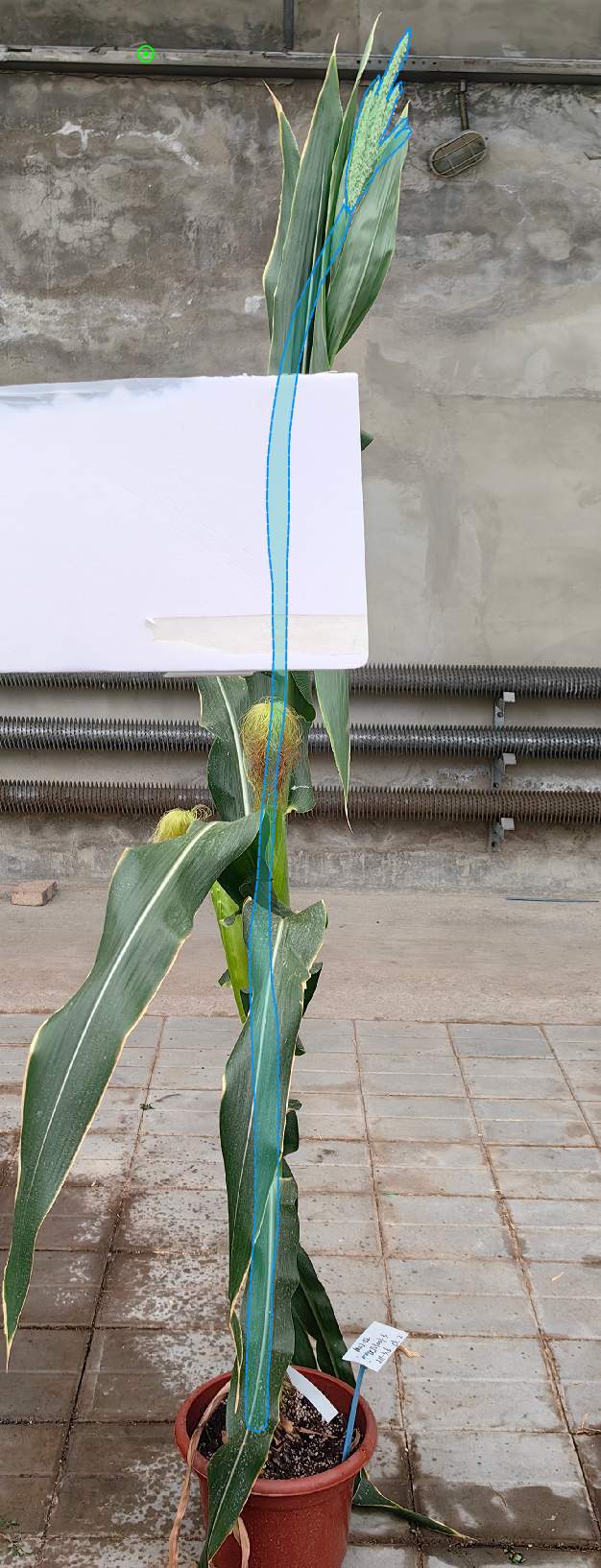} &
\includegraphics[height=0.25\textheight,keepaspectratio]{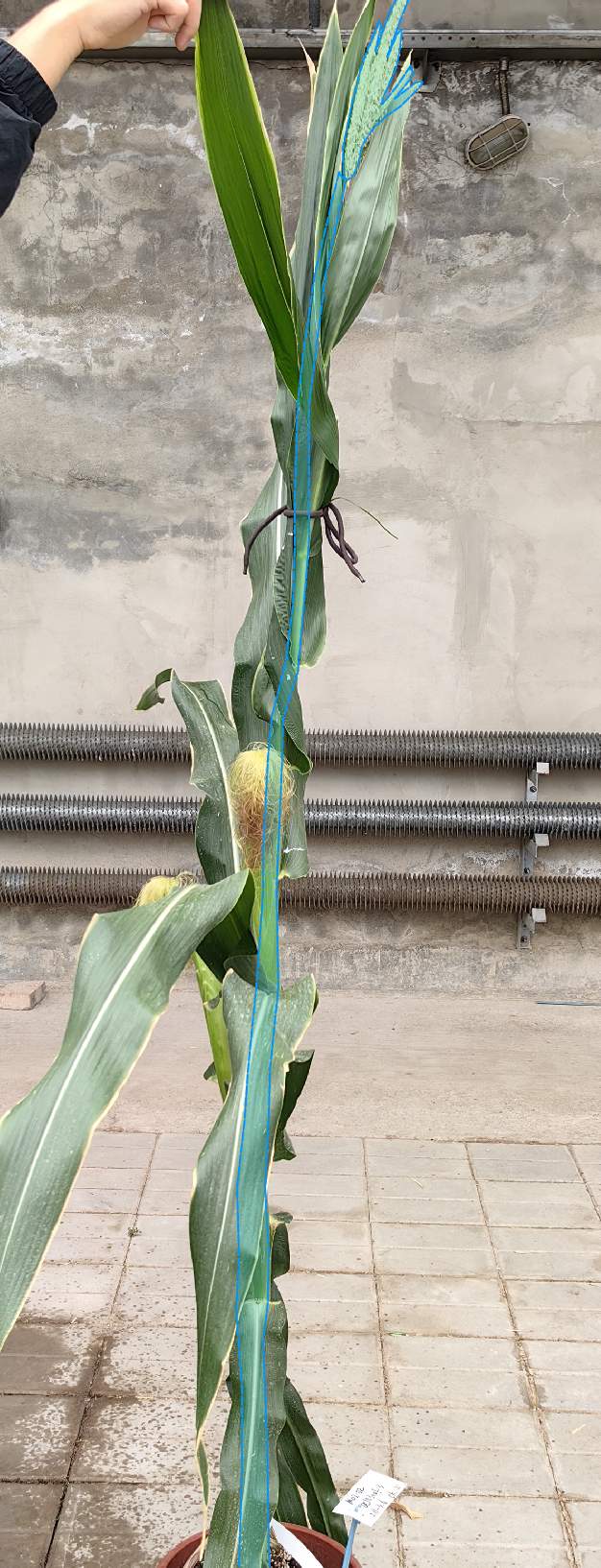} &
\includegraphics[height=0.25\textheight,keepaspectratio]{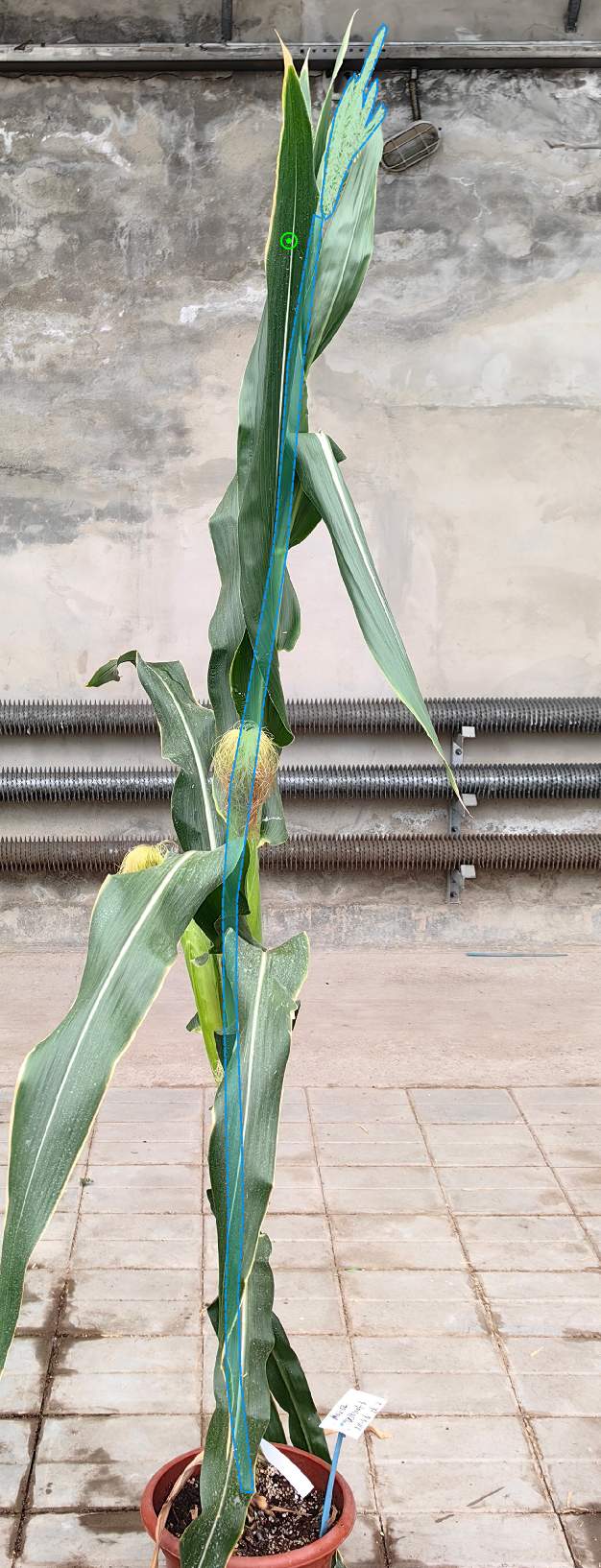} \\
{\scriptsize Full-amodal annotation} &
{\scriptsize Reference view} &
{\scriptsize Evidence-closed annotation}
\end{tabular}
\caption{
Controlled artificial-occlusion validation example.
}
\label{fig:controlled_occlusion_samples}
\end{figure}

Figure~\ref{fig:controlled_occlusion_samples} shows one representative
example from this controlled validation set. The example illustrates how
the exposed reference and occluded observations are used to compare
full-amodal and evidence-closed annotation definitions on the same
plant-level target.

This validation set is not used for model training. It is used only to
compare two annotation definitions on plant-owned stem and tassel fields:
full-amodal annotation and evidence-closed annotation. Both annotations
use the same plant-instance ownership. The exposed reference view provides
the comparison target, while the occluded views provide the annotation
inputs.

Table~\ref{tab:controlled_occlusion} reports mask mIoU and bounding-box
center distance after matching the corresponding plant and organ fields.
Higher mIoU and lower center distance indicate better agreement with the
controlled reference.

\begin{table}[t]
\centering
\small
\caption{
Controlled artificial-occlusion validation.
}
\label{tab:controlled_occlusion}
\begin{tabular}{lcc}
\toprule
Annotation definition & mIoU $\uparrow$ & BBox center dist. $\downarrow$ \\
\midrule
Full-amodal & 0.6945 & 33.72 \\
Evidence-closed & \textbf{0.8038} & \textbf{32.56} \\
\bottomrule
\end{tabular}
\end{table}

On this controlled set, evidence-closed annotation obtains higher mIoU and
a slightly lower bounding-box center distance than full-amodal annotation.
This result should be interpreted as a small-scale validation rather than
a complete proof, but it supports the practical motivation of the
EgoMaize workflow: when external occlusion is introduced, conservative
evidence-closed completion can better match the exposed reference than
unconstrained full-amodal completion.

\subsection{Inter-Annotator Consistency}
\label{sec:inter_annotator_exp}

We evaluate annotation reproducibility on a difficult subset selected
using the difficulty score defined in Section~\ref{sec:dataset_split}.
The score accounts for instance count, height-wise occlusion, and
ignore-region coverage, so higher-scoring images contain more instances,
stronger occlusion, and larger uncertain regions.

A subset of 10 images is independently annotated by four annotators. We
also evaluate full-amodal annotation under the same four-annotator
setting. For each protocol, we match corresponding plant instances across
annotators and compute pairwise mean mask IoU and bounding-box center
offset. The center offset is the Euclidean distance between the centers of
the boxes induced by paired instance masks. Table~\ref{tab:inter_annotator}
reports the resulting consistency scores.

\begin{table}[t]
\centering
\small
\caption{
Inter-annotator consistency.
}
\label{tab:inter_annotator}
\begin{tabular}{lcc}
\toprule
Protocol & mIoU $\uparrow$ & BBox center offset (px) $\downarrow$ \\
\midrule
Full-amodal & $0.5860 \pm 0.1216$ & 78.33 \\
EgoMaize & \textbf{$\mathbf{0.8203 \pm 0.0934}$} & \textbf{71.62} \\
\bottomrule
\end{tabular}
\end{table}

EgoMaize obtains a mean inter-annotator IoU of 0.8203, with a standard
deviation of 0.0934. The mean bounding-box center offset is 71.62
pixels. These results suggest that the
annotation target is reasonably reproducible despite dense overlap and
severe occlusion. The full-amodal row quantifies how much consistency is
affected when annotators are required to complete invisible projections.

\section{Conclusion}
\label{sec:conclusion}

We introduced EgoMaize, a first-person post-seedling maize instance
segmentation benchmark under severe field occlusion. The dataset targets a
practical agricultural perception problem: close-range field monitoring
and mobile phenotyping require plant-level association of stems, tassels,
and fragmented visible regions, but post-seedling maize rows contain many
same-class, elongated, and overlapping plant organs. EgoMaize provides
high-resolution RGB images, plant-level instance masks, plant-owned stem
and tassel fields, and ignore regions for unreliable maize pixels.

We also defined an evidence-closed annotation protocol for this setting.
Compared with visible-only annotation, it preserves locally supported
same-plant continuity through evidence-closed completion. Compared with
full-amodal annotation, it avoids completing hidden regions whose
ownership or boundary cannot be verified. This protocol may also be
transferable to other annotation settings where partial occlusion,
instance ownership, and unverifiable hidden regions must be handled
jointly. Benchmark results show that current instance segmentation models
can localize maize plants to some extent, but ownership-consistent mask
prediction remains difficult under elongated structures, repeated organs,
same-class overlap, and ignore regions.

Despite these contributions, EgoMaize has several limitations. It is
compact in scale and currently covers limited field sites, devices, and
acquisition conditions. The protocol still requires annotators to judge
plant ownership and evidence closure, and 2D images cannot fully validate
hidden plant geometry. The current release also focuses on plant
instances and selected organ cues rather than complete phenotyping
records.

These limitations point to several future directions. Future work will
expand the dataset across more field conditions, cultivars, planting
densities, and post-seedling growth stages. Procedural or developmental
maize models with known plant ownership and occlusion geometry could
further provide in-silico validation for hidden regions and annotator
consistency. We also plan to add phenotype-oriented attributes for exposed
stems, leaves, and tassels. The annotation schema supports incremental
annotation, so future organ traits and quality-control fields can be added
without redefining the dataset. We have also collected manual correction
traces during annotation, although they are not used in this work; future
work will use them to improve annotation quality control, active learning,
and model-assisted labeling.

\FloatBarrier

\bibliography{egbib}

\end{document}


\maketitle

\section{Data Acquisition Details}
\label{app:acquisition}

Table~\ref{tab:appendix_acquisition} summarizes the field acquisition
sessions. These details are reported for dataset documentation and are not
used as supervision labels.

\begin{table}[htbp]
\centering
\scriptsize
\setlength{\tabcolsep}{3pt}
\renewcommand{\arraystretch}{1.08}
\caption{
Field acquisition details.
}
\label{tab:appendix_acquisition}
\begin{tabularx}{\linewidth}{@{}l l c l X@{}}
\toprule
Date & Site & Images & Weather & Maize type / density (plants/mu) \\
\midrule
2026-03-25 AM & Site A & 90 & Sunny, light wind
& sweet corn / 5,500 \\

2026-03-28 PM & Site A & 61 & Sunny, light wind
& sweet corn / 5,500 \\

2026-03-29 PM & Site B & 25 & Cloudy, light wind
& grain corn / 5,000 \\

2026-03-30 & Site B & 36 & Cloudy, light wind
& waxy corn / 5,500 \\

2026-04-01 & Site C & 72 & Cloudy, light wind
& waxy corn / 5,500 \\

2026-03-28 & Site D & 17 & Cloudy, light wind
& sweet corn / 5,000 \\
\midrule
Total & -- & 301 & -- & -- \\
\bottomrule
\end{tabularx}
\end{table}

Sites A--D denote four anonymized field locations in Sanya. Device/height was
Huawei Mate 50 Pro at 160/100 cm for Sites A--C and Redmi K60 at 160 cm for
Site D. Planting density is reported in plants per mu, where one mu is
approximately 666.7 m$^2$.

\section{Annotation Tool}
\label{app:annotation_tool}

To support the EgoMaize annotation workflow, we developed a dedicated
maize annotation and pre-annotation tool. The tool supports manual mask
editing, SAM-assisted pre-annotation~\cite{kirillov2023sam}, instance
refinement, batch image management, COCO-compatible import/export, and
correction-record export. In this work, the tool is used only to improve
annotation efficiency; all annotations are manually reviewed and refined
according to the EgoMaize annotation workflow. The annotations are stored
in a COCO-style instance segmentation format, with dataset-specific
extensions for ignore regions and polygon-fragment records. The tool and
annotations will be released after acceptance.

\FloatBarrier
\bibliography{egbib}